\documentclass{article}
\usepackage{spconf,amsmath,graphicx}

\usepackage{times}
\usepackage{epsfig}
\usepackage{graphicx}
\usepackage{amsmath}
\usepackage{amssymb}
\usepackage{color}
\usepackage[numbers,sort,compress]{natbib}
\usepackage{bbm}
\usepackage[pagebackref=true,breaklinks=true,letterpaper=true,colorlinks,bookmarks=false]{hyperref}
\usepackage{epsfig}
\usepackage{enumerate,booktabs}
\usepackage{multirow}
\usepackage{subcaption}
\usepackage{setspace}

\def\x{{\mathbf x}}
\def\L{{\cal L}}

\DeclareMathOperator*{\softmax}{softmax}

\title{Knowledge Distillation for Human Action Anticipation}
\name{Vinh Tran, Yang Wang, Zekun Zhang, Minh Hoai }
\address{Stony Brook University, Stony Brook, NY 11790}
\begin{document}
\def\mA{\mathcal{A}}
\def\mB{\mathcal{B}}
\def\mC{\mathcal{C}}
\def\mD{\mathcal{D}}
\def\mE{\mathcal{E}}
\def\mF{\mathcal{F}}
\def\mG{\mathcal{G}}
\def\mH{\mathcal{H}}
\def\mI{\mathcal{I}}
\def\mJ{\mathcal{J}}
\def\mK{\mathcal{K}}
\def\mL{\mathcal{L}}
\def\mM{\mathcal{M}}
\def\mN{\mathcal{N}}
\def\mO{\mathcal{O}}
\def\mP{\mathcal{P}}
\def\mQ{\mathcal{Q}}
\def\mR{\mathcal{R}}
\def\mS{\mathcal{S}}
\def\mT{\mathcal{T}}
\def\mU{\mathcal{U}}
\def\mV{\mathcal{V}}
\def\mW{\mathcal{W}}
\def\mX{\mathcal{X}}
\def\mY{\mathcal{Y}}
\def\mZ{\mathcal{Z}}

\def\1n{\mathbf{1}_n}
\def\0{\mathbf{0}}
\def\1{\mathbf{1}}

\def\A{{\bf A}}
\def\B{{\bf B}}
\def\C{{\bf C}}
\def\D{{\bf D}}
\def\E{{\bf E}}
\def\F{{\bf F}}
\def\G{{\bf G}}
\def\H{{\bf H}}
\def\I{{\bf I}}
\def\J{{\bf J}}
\def\K{{\bf K}}
\def\L{{\bf L}}
\def\M{{\bf M}}
\def\N{{\bf N}}
\def\O{{\bf O}}
\def\P{{\bf P}}
\def\Q{{\bf Q}}
\def\R{{\bf R}}
\def\S{{\bf S}}
\def\T{{\bf T}}
\def\U{{\bf U}}
\def\V{{\bf V}}
\def\W{{\bf W}}
\def\X{{\bf X}}
\def\Y{{\bf Y}}
\def\Z{{\bf Z}}

\def\a{{\bf a}}
\def\b{{\bf b}}
\def\c{{\bf c}}
\def\d{{\bf d}}
\def\e{{\bf e}}
\def\f{{\bf f}}
\def\g{{\bf g}}
\def\h{{\bf h}}
\def\i{{\bf i}}
\def\j{{\bf j}}
\def\k{{\bf k}}
\def\l{{\bf l}}
\def\m{{\bf m}}
\def\n{{\bf n}}
\def\o{{\bf o}}
\def\p{{\bf p}}
\def\q{{\bf q}}
\def\r{{\bf r}}
\def\s{{\bf s}}
\def\t{{\bf t}}
\def\u{{\bf u}}
\def\v{{\bf v}}
\def\w{{\bf w}}
\def\x{{\bf x}}
\def\y{{\bf y}}
\def\z{{\bf z}}

\def\balpha{\mbox{\boldmath{$\alpha$}}}
\def\bbeta{\mbox{\boldmath{$\beta$}}}
\def\bdelta{\mbox{\boldmath{$\delta$}}}
\def\bgamma{\mbox{\boldmath{$\gamma$}}}
\def\blambda{\mbox{\boldmath{$\lambda$}}}
\def\bsigma{\mbox{\boldmath{$\sigma$}}}
\def\btheta{\mbox{\boldmath{$\theta$}}}
\def\bomega{\mbox{\boldmath{$\omega$}}}
\def\bxi{\mbox{\boldmath{$\xi$}}}
\def\bnu{\mbox{\boldmath{$\nu$}}}                                  
\def\bphi{\mbox{\boldmath{$\phi$}}}
\def\bmu{\mbox{\boldmath{$\mu$}}}

\def\bDelta{\mbox{\boldmath{$\Delta$}}}
\def\bOmega{\mbox{\boldmath{$\Omega$}}}
\def\bPhi{\mbox{\boldmath{$\Phi$}}}
\def\bLambda{\mbox{\boldmath{$\Lambda$}}}
\def\bSigma{\mbox{\boldmath{$\Sigma$}}}
\def\bGamma{\mbox{\boldmath{$\Gamma$}}}

\newcommand{\myminimum}[1]{\mathop{\textrm{minimum}}_{#1}}
\newcommand{\mymaximum}[1]{\mathop{\textrm{maximum}}_{#1}}    
\newcommand{\mymin}[1]{\mathop{\textrm{minimize}}_{#1}}
\newcommand{\mymax}[1]{\mathop{\textrm{maximize}}_{#1}}
\newcommand{\mymins}[1]{\mathop{\textrm{min.}}_{#1}}
\newcommand{\mymaxs}[1]{\mathop{\textrm{max.}}_{#1}}  
\newcommand{\myargmin}[1]{\mathop{\textrm{argmin}}_{#1}} 
\newcommand{\myargmax}[1]{\mathop{\textrm{argmax}}_{#1}} 
\newcommand{\myst}{\textrm{s.t. }}

\newcommand{\denselist}{\itemsep -1pt}
\newcommand{\sparselist}{\itemsep 1pt}

\definecolor{pink}{rgb}{0.9,0.5,0.5}
\definecolor{purple}{rgb}{0.5, 0.4, 0.8}   
\definecolor{gray}{rgb}{0.3, 0.3, 0.3}
\definecolor{mygreen}{rgb}{0.2, 0.6, 0.2}

\newcommand{\cyan}[1]{\textcolor{cyan}{#1}}
\newcommand{\red}[1]{\textcolor{red}{#1}}  
\newcommand{\blue}[1]{\textcolor{blue}{#1}}
\newcommand{\magenta}[1]{\textcolor{magenta}{#1}}
\newcommand{\pink}[1]{\textcolor{pink}{#1}}
\newcommand{\green}[1]{\textcolor{green}{#1}} 
\newcommand{\gray}[1]{\textcolor{gray}{#1}}    
\newcommand{\mygreen}[1]{\textcolor{mygreen}{#1}}    
\newcommand{\purple}[1]{\textcolor{purple}{#1}}       

\definecolor{greena}{rgb}{0.4, 0.5, 0.1}
\newcommand{\greena}[1]{\textcolor{greena}{#1}}

\definecolor{bluea}{rgb}{0, 0.4, 0.6}
\newcommand{\bluea}[1]{\textcolor{bluea}{#1}}
\definecolor{reda}{rgb}{0.6, 0.2, 0.1}
\newcommand{\reda}[1]{\textcolor{reda}{#1}}

\def\changemargin#1#2{\list{}{\rightmargin#2\leftmargin#1}\item[]}
\let\endchangemargin=\endlist
                                               
\newcommand{\cm}[1]{}

\newcommand{\mtodo}[1]{{\color{red}$\blacksquare$\textbf{[TODO: #1]}}}
\newcommand{\myheading}[1]{\vspace{1ex}\noindent \textbf{#1}}
\newcommand{\htimesw}[2]{\mbox{$#1$$\times$$#2$}}


\newif\ifshowsolution
\showsolutiontrue

\ifshowsolution  
\newcommand{\Comment}[1]{\paragraph{\bf $\bigstar $ COMMENT:} {\sf #1} \bigskip}
\newcommand{\Solution}[2]{\paragraph{\bf $\bigstar $ SOLUTION:} {\sf #2} }
\newcommand{\Mistake}[2]{\paragraph{\bf $\blacksquare$ COMMON MISTAKE #1:} {\sf #2} \bigskip}
\else
\newcommand{\Solution}[2]{\vspace{#1}}
\fi

\newcommand{\truefalse}{
\begin{enumerate}
	\item True
	\item False
\end{enumerate}
}

\newcommand{\yesno}{
\begin{enumerate}
	\item Yes
	\item No
\end{enumerate}
}

%
\maketitle
\begin{abstract}
We consider the task of training a neural network to anticipate human actions in video. This task is challenging given the complexity of video data, the stochastic nature of the future, and the limited amount of annotated training data. In this paper, we propose a novel knowledge distillation framework that uses an action recognition network to supervise the training of an action anticipation network, guiding the latter to attend to the relevant information needed for correctly anticipating the future actions. This framework is possible thanks to a novel loss function to account for positional shifts of semantic concepts in a dynamic video. The knowledge distillation framework is a form of self-supervised learning, and it takes advantage of unlabeled data. Experimental results on JHMDB and EPIC-KITCHENS dataset show the effectiveness of our approach.
\end{abstract}
%
%
\section{Introduction}

Human action anticipation is notoriously difficult due to the stochastic nature of the future. Given what is occurring or what can be observed in a video at the current moment, there are multiple possibilities that can happen. Thus, there is a fundamental limit to what we can anticipate, even when we have an infinite amount of training data. In practice, the amount of annotated training data is limited, so anticipation is a much harder problem. 

One common approach to address anticipation is to use supervised learning (e.g.~\cite{Hoai-DelaTorre-IJCV14,Vondrick-et-al-CVPR16,m_Wang-Hoai-FG18,m_Wang-etal-CVPR20,m_Wang-Hoai-CVIU18}), but learning a direct mapping between distant time steps can be challenging due to the weak correlation between the time steps. Suppose we are interested in anticipation with the lead time $\tau$, we can used supervised learning and train a neural network to map from the video observation \emph {up until} time $t$ (denoted $\x_t$) to the human action label $y_{t+\tau}$ at time $t + \tau$. That is to use a set of annotated training data pairs $\{\x_t, y_{t+\tau}\}$ to train a network $\mA: \x_t \rightarrow y_{t +\tau}$. 
To some extent, the training of the anticipation network $\mA$ can be done similarly to the training of a recognition network $\mR$ that maps from $\x_{t+\tau}$ to $y_{t+\tau}$, with the only difference being that the input to $\mA$ is $\x_t$ while the input to $\mR$ is $\x_{t+\tau}$.  
In general, the correlation between $\x_t$ and $y_{t+\tau}$ is weaker than the correlation between $\x_{t+\tau}$ and $y_{t + \tau}$, so the asymptotic performance of~$\mA$ is expected to be lower than the asymptotic performance of $\mR$. Furthermore, $\mA$ will converge to its asymptotic performance slower than~$\mR$. This is due to the complexity of video data, and it will take much training data to separate the relevant features from the irrelevant ones. This separation task is harder for training the anticipation network than for training the recognition network due to the higher ratio of irrelevant features. In general, it will require more training data to get the anticipation network $\mA$ to ``attend'' to the relevant features. 

\begin{figure}[t]
\begin{center}
\includegraphics[width=0.8\linewidth]{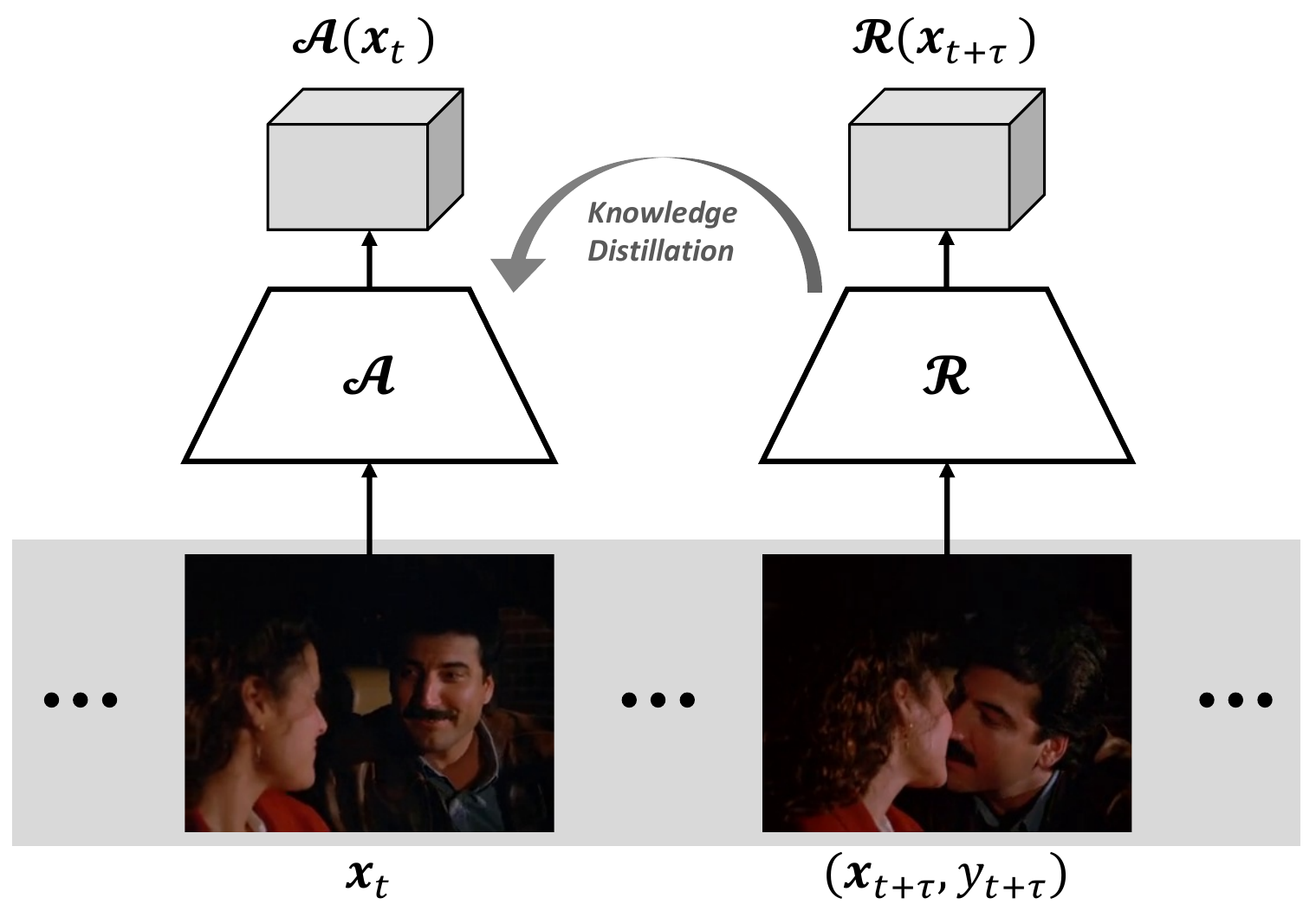}
\vspace{-0.2in}
\end{center}
   \caption{{\bf Knowledge Distillation for Action Anticipation.} 
   We propose to first learn an action recognition model $\mR: \x_{t+\tau} \rightarrow y_{t+\tau}$, then use $\mR$ to supervise the training of $\mA$ through a novel knowledge distillation framework.}
\label{fig: teaser}
\end{figure}
We propose a framework to train the anticipation network to attend to the same type of information that is being attended by the recognition network when making classification decisions. Our framework leverages the abundance of (unlabeled) data, improving the generalization ability of an anticipation network without requiring additional human annotation. However, due to dynamic environment in a video, we cannot use $L_2$ loss to force the one-to-one mapping~\cite{Vondrick-et-al-CVPR16} between elements of two activated feature maps at $t$ and $t+\tau$. Inspired by~\cite{m_Wang-Hoai-CVPR18,m_Wang-etal-BMVC20}, we propose a novel attention mechanism that does not require pixel-to-pixel correspondence between two input videos or between two feature maps. 


Experiments on three datasets show that the proposed knowledge distillation framework improves the performance of the anticipation network. The level of improvement is consistent with the level of improvement obtained as if the annotated training data is doubled. 

\begin{figure}[t]
\begin{center}
\includegraphics[width=0.8\linewidth]{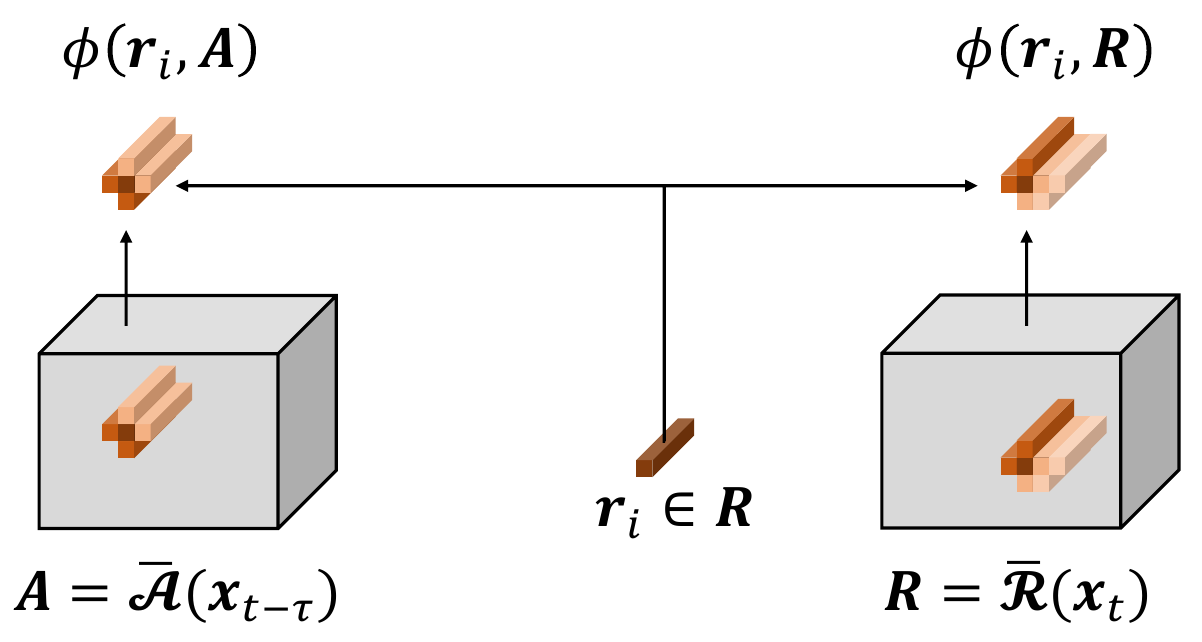}
\vspace{-0.2in}
\end{center}
   \caption{{\bf Knowledge distillation for weakly aligned feature maps.} 
   To perform knowledge distillation for weakly aligned feature maps, we propose an attentional pooling operator $\phi$ to compute the amount of $\r_i$ within $\R$ and $\A$ respectively, then minimize $ \sum_i \left\lVert \phi(\r_i, \R) - \phi(\r_i, \A) \right\rVert_2^2$.}
\label{fig: attention_pool}
\end{figure}

\section{Future Knowledge Distillation}
In this section, we describe a knowledge distillation framework that uses a recognition network to guide the training of an anticipation network, leveraging the abundance of unlabeled data. 

\subsection{Framework overview}
Suppose the desired anticipation lead time is $\tau$, our goal is to train an anticipation network $\mA$ to map from the input video segment at time $t - \tau$ (denoted $\x_{t-\tau}$) to the human action label $y_{t}$ at time $t$. That is to train $\mA$ so that $\mA(\x_{t-\tau}) = y_{t}$. We assume there is a recognition network $\mR$ to recognize the action class of an observed video clip, predicting $y_{t}$ from $\x_{t}$. 

\newcommand{\mbA}[0]{\bar{\mA}}
\newcommand{\mbR}[0]{\bar{\mR}}


Let $\mbA(\x)$ denote the feature vector/map at a particular layer of the anticipation network for the input video $\x$. Similarly, let $\mbR(\x)$ be the feature vector/map of the recognition network for the input video $\x$.
Let $\mS$ be the set of time indexes where the frames are annotated with human action labels; $t$ is in $\mS$ if $y_t$ is available. One approach for training the anticipation network is to  minimize the following classification loss defined on annotated training data as $\sum_{t \in \mS} \mL_{c}(\mA(\x_{t - \tau}), y_t).$
Here, $\mL_{c}$ is a loss function that penalizes the difference between the prediction output $\mA(\x_{t - \tau})$ and the actual class label $y_t$, e.g., using the negative log likelihood loss. 

Let $\mU$ be the set of time indexes $t$'s where $y_t$ is not available (i.e., unlabeled data). Our knowledge distillation framework optimizes the below loss function: 
\begin{align}
    & \sum_{t \in \mS} \mL_{c}(\mA(\x_{t - \tau}), y_t) +  \sum_{t \in \mU} \mL_{c}(\mA(\x_{t - \tau}), \mR(\x_{t})) \\ 
    &+ \lambda  \sum_{t \in \mU \cup \mS}  \mL_{d}(\mbA(\x_{t-\tau}), \mbR(\x_t)).
\end{align}
The above objective function trains the anticipation network $\mA$ to output the same output as the recognition network on the unlabeled data $\mU$. Furthermore, $\mL_d$ is a loss function that measures the discrepancy between two feature maps $\mbA(\x_{t-\tau})$ and $\mbR(\x_t)$. This loss trains the anticipation network to produce the same feature map as the feature map of the recognition network. This formulation uses unlabeled data and the distilled knowledge from the recognition network to guide the anticipation network to attend to the relevant information that is useful for categorizing the future action. 

\subsection{Distillation loss}
We now describe the loss function $\mL_d$ for measuring the differences between two activation feature maps. At first glance, a reasonable option for this loss function is to use the sum of squared differences between the elements of the two feature maps. However, this loss assumes perfect correspondence between the elements of the feature maps. This is too restrictive, as will be explained below.

We use convolutional architectures for the anticipation and recognition networks, and the feature maps $\mbA(\x)$ and $\mbR(\x)$ are typically 4D tensors: $\mbA(\x), \mbR(\x) \in \Re^{l{\times}h{\times}w{\times}d}$. 
Usually, $l$, $h$, and~$w$  can be obtained by dividing the length, the height, and the width of the video $\x$ by their effective convolutional strides respectively. $d$ is the number of channels of the feature map. 

Consider a particular video segment $\x_t$, the feature map $\mbR(\x_t)$ encodes the activated features important for recognizing the human action. For example, in order for the recognition network to recognize a ``wash a dish'' action, some part of the feature map might indicate the presence of the dish in the video. Arguably, for the anticipation network to successfully anticipate the ``wash a dish'' action, there must be some activated ``dish'' features in its feature map. The knowledge distillation framework encourages that by training the anticipation network to output the `same' feature map as the recognition network. 
However, it would be unreasonable to assume the ``dish'' feature to stay at the same spatiotemporal location of the feature map. More generally, video is a dynamic environment, where important objects and other semantic entities might not remain at the same locations, as illustrated in Fig.~\ref{fig: teaser}. Thus, it is unreasonable to use the sum of squared differences to measure the discrepancy between two feature maps of two different time steps.

For brevity, let us reshape the 4D tensors $\mbA(\x_{t-\tau})$ and $\mbR(\x_t)$ to 2D matrices $\A$ and $\R$, $\A = [\a_1, \ldots, \a_{lhw}] \in \Re^{d{\times}lhw}$, $\R = [\r_1, \ldots, \r_{lhw}] \in \Re^{d{\times}lhw}$. 
One naive approach is to directly minimize $\mL_{direct}(\A, \R) = \left\lVert \A - \R\right\rVert_2^2.$
However, it would be unreasonable because the features in $\A$ and $\R$ might be at different spatiotemporal locations after $\tau$ seconds.
Instead, for each vector~$\a_i$, we measure the similarity between $\a_i$ with all vectors in $\R$ and we compute a vector quantity to represent the amount of $\a_i$ in $\R$: $\phi(\a_i, \R) = \sum_{k=1}^{lhw} \omega_k\r_k$,
with $\omega_k = \frac{1}{Z}\exp(\alpha \r_k^T\a_i)$, where $Z$ is the normalizing constant so that the sum of $\omega_k$'s is 1. Here, $\alpha$ is the hyper-parameter that controls the pooling weights. The default value of $\alpha$ is set to $\frac{1}{\sqrt{d}}$, where $d$ is the number of channels of the feature maps. If the value of~$\alpha$ is small, the weights associated to each vector are almost equal, and $\phi(\a_i, \R)$ is the average pooling of vectors in $\R$. On the other hand, this operator is similar to max pooling if we use a large value for $\alpha$, and $\phi(\a_i, \R)$ is the vector in $\R$ that is most similar to $\a_i$.  Equivalently, $\phi(\a_i, \R)$ and $\phi(\a_i, \A)$ can be expressed in the form: 
$\phi(\a_i, \R) = \R\softmax(\alpha \R^T\a_i)$, and $\phi(\a_i, \A) = \A\softmax(\alpha \A^T\a_i)$. We define the loss for the differences between two feature maps $\A$ and $\R$ as follows:
\begin{align}
    &\mL_d(\A, \R) =  \tilde{\mL}_d(\A, \R) + \tilde{\mL}_d(\R, \A),  \label{equ: distill_loss} \\
\textrm{where } \ \
    &\tilde{\mL}_d(\A, \R) = \sum_{i=1}^{lhw} || \phi(\a_i, \R) - \phi(\a_i, \A)||_2^2,  \\
    &\tilde{\mL}_d(\R, \A) = \sum_{i=1}^{lhw} || \phi(\r_i, \A) - \phi(\r_i, \R)||_2^2. 
\end{align}

\section{Experiments}
\subsection{Datasets}
We conducted the main experiments on two challenging datasets: JHMDB~\cite{jhuang2013towards} and  EPIC-KITCHENS~\cite{Damen2018EPICKITCHENS}. We also performed some controlled experiments on the THUMOS dataset \cite{Gorban-etal_THUMOS15} to understand the expected benefits of having extra supervision.

\subsection{Experiments on the JHMDB dataset}
We performed several experiments on the JHMDB dataset. We used the I3D network as the backbone for this task. We followed the standard protocol~\cite{shi2018action} for evaluation on this dataset, i.e., using only the first 20\% of the frames to predict action class labels.
During training, we combined both the classification loss $\mL_c$ (using the class labels or the predicted class probability) and the attention loss $\mL_d$ (using feature maps). We used KL divergence for the classification loss. For the attention loss, we used Huber loss (with $\delta=1$). 
Both RGB frames and optical flow maps were used to train an anticipation network.

\begin{table}[t]
\begin{center}
\small
\begin{tabular}{lcccc}
\toprule
Method & RGB & Flow & Both\\
\midrule
\textbf{Recognition Network} \\
\hspace{3mm}I3D  & 75.3 &    77.8 &    83.9 \\
\midrule
\textbf{Anticipation Network} \\
\hspace{3mm}I3D                              & 69.0 & 64.4 & 74.9 \\
\hspace{3mm}I3D + ${\mL}_{direct}$           & 69.5 & 67.1 & 75.5 \\
\hspace{3mm}I3D + $\tilde{\mL}_{d}(\R, \A)$  & 70.0 & 67.4 & 75.0 \\
\hspace{3mm}I3D + $\tilde{\mL}_{d}(\A, \R)$  & 69.5 & 67.5 & 75.8 \\
\hspace{3mm}I3D + $\mL_{d}(\A, \R)$  & \textbf{70.2} &    \textbf{67.7} & \textbf{76.6} \\
\bottomrule
\end{tabular}
\end{center}
\vspace{-0.2in}
\caption{{\bf Action anticipation results on the JHMDB dataset.} 
The recognition network uses the entire video for classification while the anticipation network only observes the first 20\% of the video. } \label{table: jhmdb}

\end{table}

\begin{table}[t]
\begin{center}
\small
\begin{tabular}{lc}
\toprule
Method &  Acc(\%)\\
\midrule
Where/What~\cite{soomro2016predicting} & 10.0 \\
Context-fusion~\cite{jain2016recurrent} & 28.0 \\
Within-class Loss~\cite{Ma-et-al-CVPR16} & 33.0\\
ELSTM~\cite{sadegh2017encouraging} & 55.0\\
FDI~\cite{RodriguezEccv18} & 61.0 \\
FM-RNN~\cite{shi2018action} & 73.4 \\
\midrule
I3D + Knowledge Distillation (Ours)  & \textbf{76.6} \\
\bottomrule
\end{tabular}
\end{center}
\vspace{-0.2in}
\caption{{\bf Comparison of action anticipation methods on the JHMDB dataset.} 
All methods use the first 20\% of the video for prediction. } 
\label{table: jhmdb_soa}
\end{table}

We report the action recognition and anticipation performance of different methods in Tab.~\ref{table: jhmdb}.
First, we trained the action recognition network using all the available frames in the training set. 
Directly applying the recognition network on the first $20\%$ frames of the test videos (i.e., using the recognition network for the anticipation task), the accuracy dropped drastically to $74.9\%$. Second, we applied the direct loss (denoted as $\mL_{direct}$ as in Tab.~\ref{table: jhmdb}) between two feature maps produced by anticipation network and recognition network. With this additional loss, the accuracy was increased to $75.5\%$. This was possibly thanks to the small displacement between two feature maps since the dataset contains only action and the $15$ frames anticipation is short. Hence, the $\mL_{direct}$ loss also helped improving the recognition performance on this dataset. Replacing the direct loss function with our distillation loss, the performance increased to $75.8\%$. Finally, we achieved the best performance of $76.6\%$ when using the symmetric bidirectional attention loss $\mL_d(A,R)$. As shown in Tab.~\ref{table: jhmdb_soa}, we obtained the new state of the art result on the JHMDB dataset.

\subsection{Experiments on the Epic-Kitchens dataset}

\begin{table}[t]
\begin{center}
\begin{tabular}{lc}
\toprule
Method &  Acc.(\%) \\
\midrule
R(2+1)D + Vis. Attr.~\cite{Miech-et-al-CVPR19} & 28.4 \\
TSN-RGB~\cite{TSN2016ECCV} & 28.5\\
TSM-RGB~\cite{Lin-ICCV19-TSM} & 30.3 \\
I3D~\cite{carreira2017i3d} & 30.1 \\
\midrule
I3D + Data augmentation & 29.8 \\
I3D + Additional data only (Ours)  & 31.4\\
I3D + Knowledge Distillation (Ours)  & \textbf{31.8}\\
\bottomrule
\end{tabular}
\vspace{-0.2in}
\end{center}
\caption{{\bf Accuracy of anticipation methods on the EPIC-KITCHENS dataset} (for anticipating the verb actions). All methods reported here are implemented by us, trained with the same amount of labeled data. }
\label{table: epic_verb}
\vspace{-0.1in}
\end{table}

We used the I3D network architecture for the experiments described in this subsection, as in the previous subsection. 
Since Epic-Kitchen is a large dataset, we used the feature maps extracted from the \texttt{MaxPool3d\_4a\_3x3} layer as the input to the network instead of training directly from video frames. 

\myheading{Collecting unlabeled training data.}
We collected video clips from unlabeled segments as follows. First, we randomly took two video segments of $32$ frames with the anticipation time $\tau{=}1$s from unlabeled video segments. Second, we used the pre-trained I3D  to extract feature maps at the \texttt{MaxPool3d\_4a\_3x3} layer for the two video segments. Third, we fed the feature map of the latter segment to the recognition network and computed its visual representation (i.e., both class probabilities and feature maps). Finally, the feature map of the first segment and the visual representation of the second segment formed a data-pair sample for training the anticipation network. We collected a total of $26,391$ unlabeled training samples. Together with the $23,191$ labeled examples, we had a total of $49,582$ training samples. This is roughly double the amount of the original training data. 

\myheading{Experimental Results}.
We report all performance values in Tab.~\ref{table: epic_verb}. 
When we trained a I3D network using the annotated training data, we achieved an accuracy of $30.1\%$. This was better than the performance of the TSN-RGB~\cite{TSN2016ECCV}, which achieved $28.5\%$ accuracy on the same dataset. We also trained another I3D network with augmented training data, where we used both the video segments prior to the actions and the video segments right at the time of the actions to train the network. However, this approach slightly decreased the performance of the anticipation network. This was perhaps due to the use of `noisy data', since the video segments at the time of the actions were meant for the recognition network not the anticipation network. 
Using additional unlabeled data with pseudo label, we improved the accuracy to $31.4$.
Using knowledge distillation and additional unlabeled data, the obtained anticipation network obtained $1.7\%$  improvement in accuracy ($30.1 \rightarrow {31.8}$). This was also better than the recent video network TSM-RGB~\cite{Lin-ICCV19-TSM} method ($30.3\%$).

\begin{figure}[t]
\begin{center}
\includegraphics[width=0.8\linewidth]{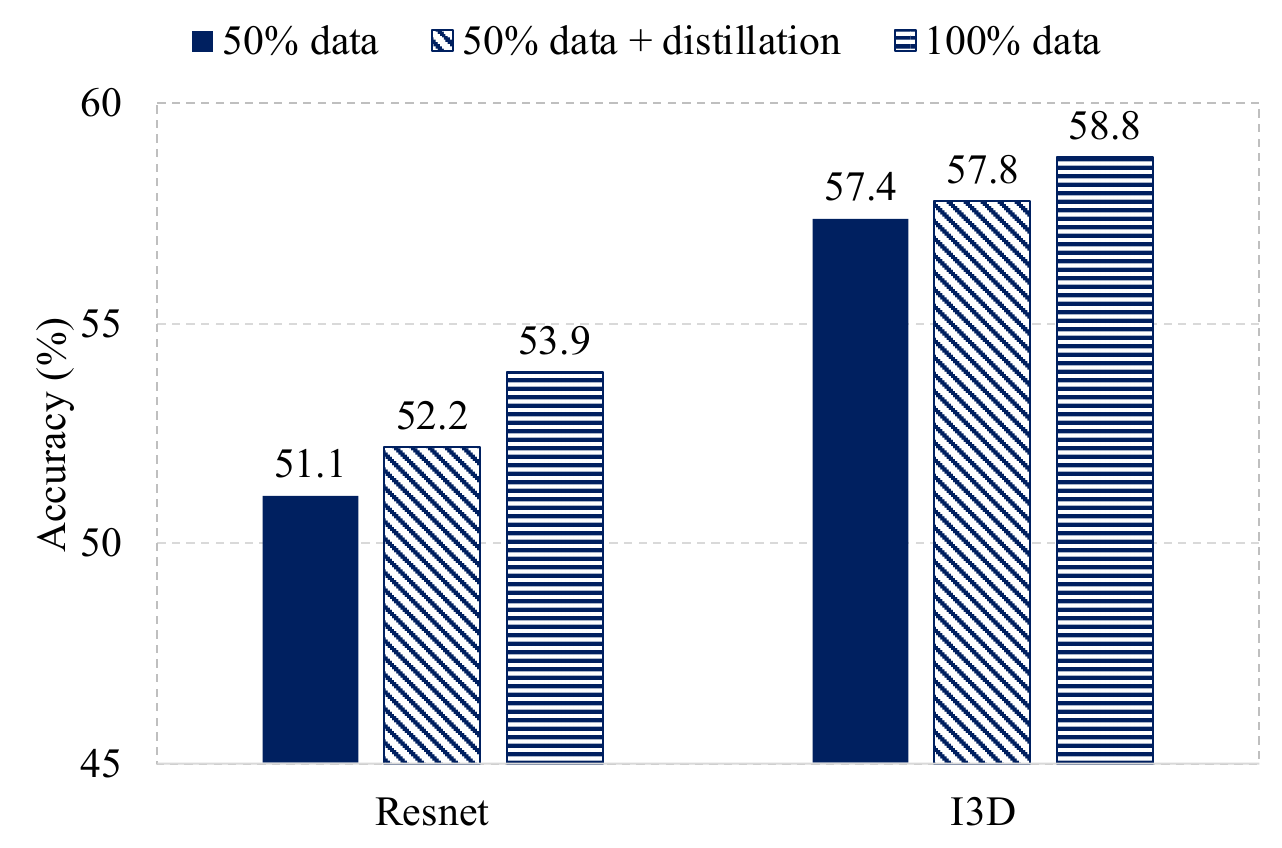}
\vspace{-0.25in}
\end{center}
  \caption{{\bf Expected performance gain when doubling the amount of training data.} 
  }
\label{fig: thumos14}
\vspace{-0.1in}
\end{figure}

\subsection{Experiments on THUMOS14 dataset}
We performed controlled experiments to rectify the expected level of improvement of distills knowledge from a recognition network. 
We used videos from the THUMOS14 action detection challenge~\cite{Gorban-etal_THUMOS15} to create a dataset for action anticipation. We first identified the temporal location of an action segment. Interested in the anticipation lead time of one second, we moved back one second and extracted a 1s clip ending at that location and used as the input to the anticipation network. Using this strategy, we can compile a training dataset of multiple 1s clips. We then trained two anticipation networks, one using the full training set and the other using the smaller training set with 50\% of the data. We experimented with both 2D and 3D ConvNet architectures to see how the size of the training set affects the anticipation performance.

The anticipation performance of these networks is plotted in Fig.~\ref{fig: thumos14}. As can be seen, more data improved the performance of an anticipation network.  When doubling the amount of annotated training data, the gain in accuracy of the two networks (for two types of features) were 1.4\% and 2.8\%. This experiment showed that doubling the amount of annotated training data would only yield moderate improvement in anticipation accuracy, perhaps somewhere from 1\% to 3\%.

Other important factors are the accuracy of the recognition network and also the quality of the unlabeled data. Fig.~\ref{fig: thumos14} shows the performance of the anticipation networks trained with knowledge distillation. In this experiment, we only used half of the labeled training data, and the other half as unlabeled data for knowledge distillation. 
 As can be seen, the level of improvement was not as good as having actual ground truth annotations, and this can probably be attributed to the imperfection of the recognition network. 

\section{Summary}
We have presented a framework for knowledge distillation. This framework uses the action recognition network to supervise the training of an action anticipation network. With a novel knowledge distillation technique to account for the positional drift of semantic concepts in video, the action recognition network acts as a teacher guiding the anticipation network to attend to the relevant information needed for predicting the future action. Using this framework, we are able to leverage unlabeled data to train the anticipation network in a self-supervised manner. The experimental results on the JHMDB and EPIC-KITCHENS datasets show the benefits of our proposed method.

\myheading{Acknowledgement.} {\small  This material is based on research sponsored by the Air Force Research Laboratory (AFRL), DARPA, under agreement number FA8750-19-2-1003. The U.S. Government is authorized to reproduce and distribute reprints for Governmental purposes notwithstanding any copyright notation thereon. The views and conclusions contained herein are those of the authors and should not be interpreted as necessarily representing the official policies or endorsements, either expressed or implied, of the AFRL, DARPA, or the U.S. Government.  }

\newpage

\bibliographystyle{IEEEbib}
\bibliography{shortstrings,egbib,pubs}

\end{document}